\documentclass[twocolumn, switch]{article}

\usepackage{arxiv}

\usepackage[utf8]{inputenc} 
\usepackage[T1]{fontenc}    
\usepackage{hyperref}       
\usepackage{url}            
\usepackage{booktabs}       
\usepackage{amsfonts}       
\usepackage{nicefrac}       
\usepackage{microtype}      
\usepackage{lipsum}
\usepackage[binary-units=true]{siunitx}
\usepackage{comment}
\usepackage{stfloats}
\usepackage{graphicx}
\usepackage{multirow}
\usepackage{makecell}
\usepackage{epsfig}
\usepackage{pifont}
\usepackage{booktabs} 
\usepackage{footnote}
\usepackage[english]{babel}
\usepackage{algorithmicx, algorithm, algpseudocode}

\algrenewcommand\textproc{}
\usepackage{amsmath}
\usepackage{bm}
\usepackage{multicol}
\usepackage{amssymb}
\usepackage{setspace}
\usepackage{tikz}
\usepackage{stfloats}
\usepackage{subfig}
\usepackage{regexpatch}
\usepackage{lettrine}

\newcommand\copyrighttext{
 \footnotesize \textbf{IEEE Copyright Notice.}
\textcopyright This work has been submitted to the IEEE for possible publication. Copyright may be transferred without notice, after which this version may no longer be accessible.}
\newcommand\copyrightnotice{
\begin{tikzpicture}[remember picture,overlay]
\node[anchor=south,yshift=30pt] at (current 
page.south) {\fbox{\parbox{\dimexpr\textwidth-\fboxsep-\fboxrule\relax}{\copyrighttext}}};
\end{tikzpicture}
}

\title{Dynamic ConvNets on Tiny Devices\\via Nested Sparsity}
\date{\vspace{-5ex}}
\author{
Matteo Grimaldi, Luca Mocerino, Antonio Cipolletta, Andrea Calimera\\
\texttt{ \{name.surname\}@polito.it } \\
Politecnico di Torino, 10129 Torino, Italy\vspace*{0mm}
}

\begin{document}
\twocolumn[ 
  \begin{@twocolumnfalse} 
\maketitle
\copyrightnotice 
\vspace{-0.8cm}

\begin{abstract}
This work introduces a new training and compression pipeline to build Nested Sparse ConvNets, a class of dynamic Convolutional Neural Networks (ConvNets) suited for inference tasks deployed on resource-constrained devices at the edge of the Internet-of-Things.  
A Nested Sparse ConvNet consists of a single ConvNet architecture containing $N$ sparse sub-networks with nested weights subsets, like a {\em Matryoshka} doll, and can trade accuracy for latency at run time, using the model sparsity as a dynamic knob. To attain high accuracy at training time, we propose a gradient masking technique that optimally routes the learning signals across the nested weights subsets. To minimize the storage footprint and efficiently process the obtained models at inference time, we introduce a new sparse matrix compression format with dedicated compute kernels that fruitfully exploit the characteristic of the nested weights subsets. Tested on image classification and object detection tasks on an off-the-shelf ARM-M7 Micro Controller Unit (MCU), Nested Sparse ConvNets outperform variable-latency solutions naively built assembling single sparse models trained as stand-alone instances, achieving ($i$) comparable accuracy, ($ii$) remarkable storage savings, and ($iii$) high performance. Moreover, when compared to state-of-the-art dynamic strategies, like dynamic pruning and layer width scaling, Nested Sparse ConvNets turn out to be Pareto optimal in the accuracy vs. latency space. 
\end{abstract}
\vspace{0.1cm}

\vspace{0.35cm}

\vspace{0.35cm}
\end{@twocolumnfalse}]
\section{Introduction}~\label{sec:introduction}
\vspace{-0.5cm}
\lettrine{T}{he} ability to deploy fast Convolutional Neural Networks (ConvNets) at the edge of the Internet-of-Things (IoT) reflects the possibility of building ubiquitous intelligent services with high efficiency and privacy standards. 
In many IoT applications, the end-nodes are lightweight devices powered by tiny Micro Controller Units (MCUs), characterized by small form factor, minimal storage and memory resources, i.e., few MBs of FLASH ($1$-$2$MB) and hundreds of KBs of RAM (${\leq}512kB$), and single-core CPUs clocked at few hundreds of MHz ($100$-$400$ MHz).
To bridge the gap between such stringent hardware constraints and the computational and storage requirements of modern ConvNets, a considerable research effort has been lately spent seeking optimization strategies, like pruning~\cite{he2017channel, grimaldi2020east}, precision scaling~\cite{Jacob_2018_CVPR}, compact neural architectures~\cite{howard2017mobilenets, sandler2018mobilenetv2}, and computational graph rewritings~\cite{9473965, MLSYS2020_9bf31c7f}. 
Despite the remarkable results achieved, those solutions follow a worst-case, accuracy-driven design and optimization strategy generating static ConvNets tailored for a specific setting. Static ConvNets show one main limitation, that is, they spend the same maximal effort in all situations, neglecting run-time changes that might appear due to variations in the external environmental conditions, the quality-of-service required by the user and the surrounding context, and the resources consumed by other software routines running in parallel on the same device. A speculative and perhaps more efficient approach would exploit contextual optimizations to minimize the average resource usage improving the information-processing capability. 
For instance, a video surveillance system can reduce the classification effort when the scene is empty, lowering the energy consumption, then can switch into a more accurate but expensive mode only if something suspicious is detected. Alternatively, the latency of the inference task might change to meet different throughput requirements when the resource budget at the system or the application level gets reallocated~\cite{mocerino2020fast}. These practical examples suggest that the availability of {\em dynamic} ConvNets capable of trading accuracy and computational costs at run time represents a valuable tool to raise the bar of efficiency for intelligent edge applications.

Building a dynamic ConvNet encompasses the choice of a proper control mechanism to implement the latency-quality scaling at run time.
Recent works proposed several architectural-level knobs, like the network depth~\cite{wang2018skipnet}, or the layers width~\cite{slimmable}, but although the relative ease of implementation, operating on the architecture of the model may be a too coarse option limiting the latency and accuracy trade-off.
Moreover, it does not alleviate the pressure on the storage memory as the full model configuration, i.e., the one at the maximum width or maximum depth might still be too large to fit into the FLASH memory. 
The availability of more fine-grain control knobs to modulate latency while keeping model footprint minimal is highly desirable indeed, and model {\em sparsity} is a good knob candidate. 
Sparse training is less prone to accuracy losses, and sparse models can be compressed via sparse encoding formats~\cite{hoefler2021sparsity}. 
However, how to leverage the weight sparsity as the dynamic knob on compact ConvNets, e.g., the MobileNets~\cite{howard2017mobilenets}, and how to deploy dynamic sparse ConvNets efficiently on tiny general-purpose cores are open issues.

\begin{figure}[t]
    \center
    \includegraphics[width=0.95\columnwidth]{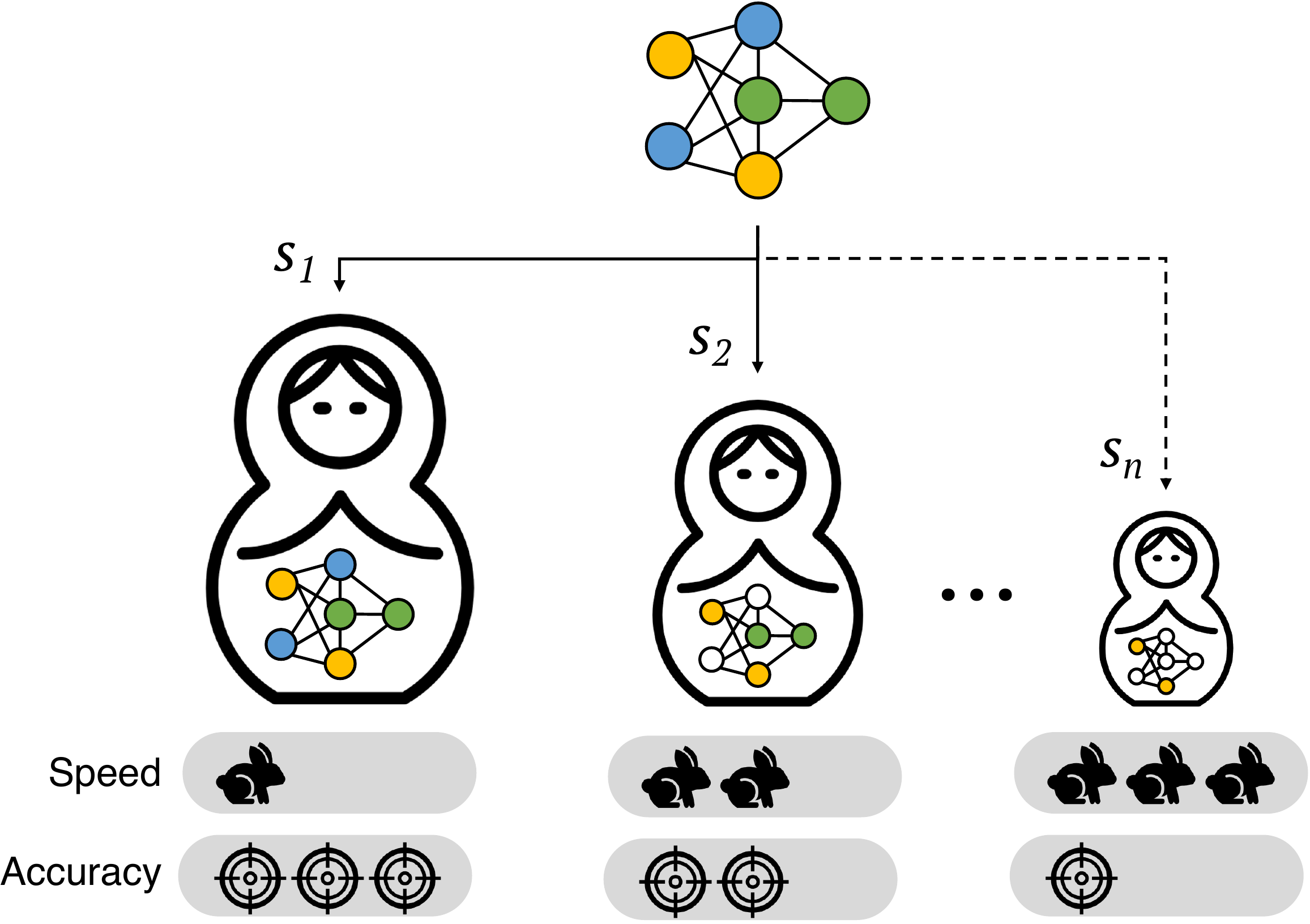}
    \vspace{2mm}
    \caption{A pictorial representation of a {\em Nested Sparse ConvNet}, a super-network containing $N$ sub-networks with increasing value of sparsity ($s_1{<}s_2{<}...{<}s_N$): a low sparsity value corresponds to high accuracy, whereas a high sparsity value results in a faster inference process at the cost of lower accuracy.
    \label{fig:matryoshka}}
\end{figure}

To this end, we introduce a new class of dynamic ConvNets named {\em Nested Sparse ConvNets}. A Nested Sparse ConvNet is a convolutional deep neural network with a single weight-set that can be operated at $N$ different configurations of increasing sparsity, resulting in a super-network containing $N$ sparse sub-networks with nested weight-sets, like {\em Matryoshka} dolls as illustrated in Fig.~\ref{fig:matryoshka}. A low sparsity value corresponds to high accuracy, whereas a high sparsity value results in a fast yet less accurate inference. 
To let any ConvNet be transformed into a Nested Sparse ConvNet, this work proposes an end-to-end pipeline that comprises three main tools integrated over the full development stack:
\begin{itemize}
\item at training time, a gradient masking technique that properly routes the learning signals between the nested sparse networks guaranteeing convergence and high accuracy;
\item at compile time, a sparse matrix compression format to fruitfully exploit the nested structure of the weights set avoiding computationally expensive decoding stages;
\item at run time, dedicated compute kernels that ensure efficient processing and switching among different sparse configurations with no additional latency cost.
\end{itemize}

To validate our proposal, we collected an extensive set of results using as benchmarks {\em ResNet9}~\cite{resnet} and two instances of {\em MobileNet} (V1 and V2)~\cite{howard2017mobilenets, sandler2018mobilenetv2} for two vision tasks, namely, image classification and object detection, deployed on an embedded system powered by an ARM Cortex-M7 MCU with $2MB$ of FLASH and $512KB$ of RAM. As it will be discussed in the experimental section, Nested Sparse ConvNets achieve an accuracy comparable to that of independently trained sparse models and outperform other scalable ConvNets obtained through existing dynamic methods, like dynamic pruning~\cite{wu2021dynamic} and layer width scaling~\cite{slimmable}, thereby proving to be Pareto optimal in the accuracy vs. latency space.

The remainder of the paper is organized as follows. Section~\ref{sec:related} reviews existing approaches to implement latency-quality scaling in ConvNets. Section~\ref{sec:methodolgy} describes the proposed end-to-end pipeline consisting of the training methodology, the compression schema, and the sparse computational kernels. Section~\ref{sec:results} presents the collected experimental results through an extensive assessment of functional and extra-functional metrics. Section~\ref{sec:limitations} discusses limitations and future works. Section~\ref{sec:conlusions} concludes the work.
\section{Related Works}~\label{sec:related}
\noindent
\hspace{-1mm}This section offers a brief review of recent works on pruning strategies and compressed sparse storage formats for static ConvNets, as the proposed pipeline extends such techniques in a dynamic context. Then it describes state-of-the-art solutions for dynamic ConvNets and the limitations that our proposal aims to overcome. 

\begin{figure*}[t]
    \center
    \includegraphics[width=.95\textwidth]{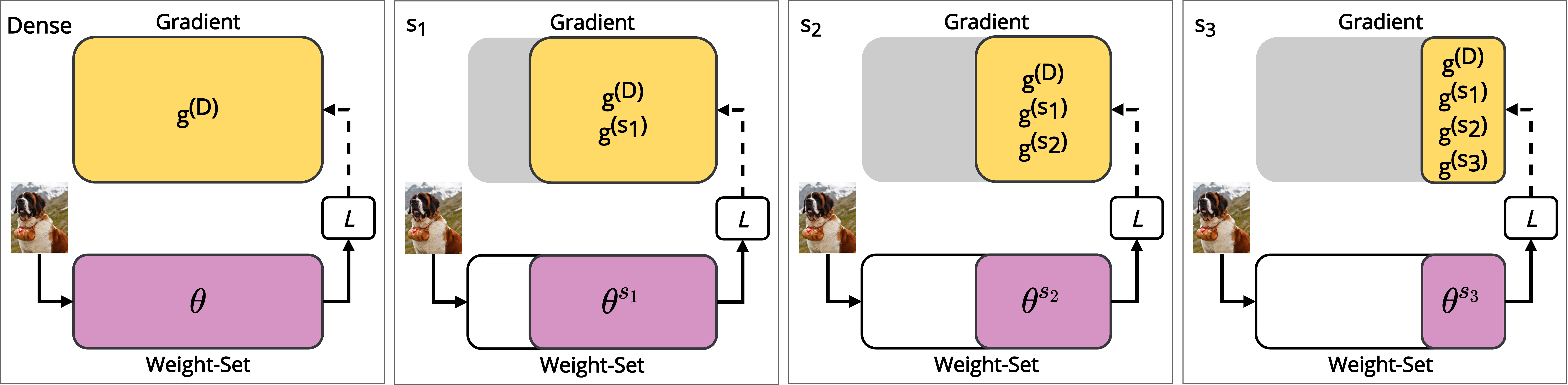}
    \vspace{2mm}
    \caption{Evolution of the training loop, from left to right. The full weight-set ($\theta$) and the sub-nets ($\theta^{s_i}$) get sorted and processed with an increasing order of sparsity value (i.e., $s_1 < s_2 <s_3$). }
    \label{fig:grad_mask}
\end{figure*}

{\bf Pruning.} The existing methods differ in terms of the pruning policy they implement and the level of granularity at which they are applied~\cite{state_sparsity}. In terms of policy, even if complex and rather elegant methods have been recently proposed, e.g., gradient- or Hebbian-based methods~\cite{hoefler2021sparsity}, those magnitude-based~\cite{han2015learning} are the preferred option in many modern training pipelines because of their reliability and ease of use. 
For what concerns the granularity, there exist three main classes. The \textit{unstructured} pruning plays at the lower level, namely, on the individual weights of the model~\cite{he2016deep}, offering a high degree of flexibility in reaching high accuracy targets. Such flexibility is paid at inference time when the potential savings brought by zeroed weights contrast with the regular code organization and memory access pattern preferred by common Instruction-Set Architectures. This issue is often solved with the aid of specialized hardware units that can accelerate the irregular flow, e.g.,~\cite{parashar2017scnn}. At a coarser granularity, \textit{block} pruning techniques~\cite{kalchbrenner2018efficient} group neighboring weights in specific patterns to decrease the indexing overhead and to ease the adoption of sparse compute kernels on general-purpose cores~\cite{scalpel, elsen2020fast}. At the coarsest level, \textit{filter} pruning schemes drop entire convolutional filters~\cite{he2017channel}, achieving aggressive storage savings and speed-up at the cost of substantial accuracy loss due to fast information removal.

{\bf Compressed Sparse Storage Format.} Dealing with sparse arrays obtained by pruning irrelevant weights enables substantial memory savings by storing the value and the position of the remaining non-zero entries. Many different sparse storage formats exist in literature~\cite{pooch1973survey} and their optimality is a function of the sparsity itself and the access pattern needed, e.g., random, streaming, or transposed access. For instance, to maximize the compression efficiency, a simple bitmap is preferable for low sparsity regimes, whereas coordinate-offset schemes (COO) are more suitable in high-sparse regimes~\cite{hoefler2021sparsity}. Sparse storage formats like Compressed Sparse Row (CSR) or Columns (CSC)~\cite{scalpel} allow fast row access, and so can be used to implement efficient sparse-matrix-vector and sparse-matrix-matrix operations.

{\bf Dynamic Topology.} One way of building scalable ConvNets is to play with the architectural structure of the model, e.g., the width of the layers or the depth of the network.
The authors of~\cite{howard2017mobilenets} proposed to scale the number of channels within each convolutional layer by a predefined ratio, the {\em width multiplier}. Originally proposed as a static design option, the authors of~\cite{slimmable} introduced the {\em switchable} batch-norm concept enabling a reliable training procedure for dynamic width scaling. Alternatively, the number of layers traversed during the forward pass can be modulated by attention modules or gating blocks~\cite{wang2018skipnet} enabling a dynamic routing of the inner features, eventually with the addition of early-exit branches~\cite{branch}. Notice that the total storage space is dictated by the underlying full-width model, or the full-depth model, plus the extra modules possibly needed for controlling the topology at run time.

{\bf Dynamic Sparsity.} Relying on the intuitive principle {\em the higher the sparsity, the lower the latency}, the authors of~\cite{wu2021dynamic} proposed a training flow for deep neural models learned under concurrent sparsity levels. Despite the preliminary results conducted on Recurrent Neural Networks (RNNs) for Automatic Speech Recognition (ASR), known to be redundant and hence more reliable to pruning~\cite{narang2017exploring}, we observed a substantial accuracy degradation on compact ConvNets for image classification tasks. Moreover, the training loop proposed in~\cite{wu2021dynamic} is unaware of the resource usage and the achievable performance, leaving the minimization of the storage footprint and the deployment on real processing cores unsolved. Our proposal addresses both issues, offering Nested Sparse ConvNets as an effective solution for ConvNet architectures deployed on actual compute nodes for the IoT.

\section{Building Nested Sparse ConvNets}~\label{sec:methodolgy}
\vspace{-1cm}
\subsection{Training} \label{sec:training}
Training a Nested Sparse ConvNet is like concurrently learning $N$ sub-networks with increasing sparsity encapsulated within a single set of weights $\theta$. Collecting and composing the learning contributions coming from (and directed to) the many sparse sub-networks is a challenging problem as the learning of the weights shared among multiple sub-networks must be properly balanced to avoid sudden accuracy drops. 
For a better understanding, let's recall how pruning techniques for static ConvNets actually work. Early methods, e.g., \cite{han2015learning}, suggested that pruned weights must be bypassed during the gradient updates, but most recent works~\cite{state_sparsity} introduced an improved pruning-while-training strategy that \textit{regrow} lost connections achieving higher accuracy results. This is the starting point for our proposal. 
Managing the \textit{regrowth} mechanism for a Nested Sparse ConvNet is not straightforward as the current ``state'' of a single weight (i.e., pruned or not-pruned) might differ among the $N$ sub-networks, generating conflicts that may prevent convergence. To handle these constraints that may bubble up during training, we developed a novel method, referred to as {\em gradient masking}, precisely conceived to route the learning signals among the sub-networks.

\begin{algorithm}
\renewcommand{\thealgorithm}{}
\caption{Nested Sparse Training}
\small
\begin{algorithmic}[1]
    \Function{main}{steps, S, block\_shape, optimizer}\label{alg:training}
    \For{t {\bf in} steps}
        \State optimizer.zero\_grad() \Comment{$\hat{G}$ = $0$}
        \State soft\_labels = forward($\theta$)
        \State $\hat{G}$ += backward($\theta$)
        \If{pruneStep(t)}
            \For{s {\bf in} S}
                \State $M^s$ = getMask($\theta$, s, block\_shape)
                \State $\theta^s = \theta \circ M^s$
                \State forward($\theta^s$, soft\_labels)
                \State $\hat{g}^{s}$ = backward($\theta^s$)
                \State $\hat{G}$ += $M^s \circ \hat{g}^{s}$ \Comment{$\hat{G}$ masking}
            \EndFor
        \EndIf
        \State optimizer.step() \Comment{$\hat{G}$ update}        
    \EndFor
    \State $M$ = \{getMask($\theta$, s, block\_shape) {\bf for} s {\bf in} S\}
    \State \Return $\theta$, $M$
    \EndFunction
    \item[]
    \Function{getMask}{$\theta$, s, block\_shape}
        \State blocks = groupBlocks($\theta$, block\_shape)
        \State idx = rankBlocks(blocks, s)
        \State mask = ones\_like($\theta$) 
        \State mask[idx] = 0
        \State \Return mask
    \EndFunction
\end{algorithmic}
\end{algorithm}

An abstract and pictorial view of the dynamics governing the training steps of a Nested Sparse ConvNet using \textit{gradient masking} is reported in Fig.~\ref{fig:grad_mask}. The example is for $N$=3 sub-networks of increasing sparsity $s_1<s_2<s_3$ and illustrates the run of a single training step. The three sub-networks are evaluated in sequence, following an increasing order of sparsity, from low ($s_1$) to high ($s_3$), as depicted within the three frames labeled as $s_1$, $s_2$, $s_3$. 
The first frame on the left (labeled as Dense) is for the full weight-set $\theta$ (i.e., sparsity $s_0{=}0\%$). The dense training ensures stability, but the dense network is not included in the final model deployed for inference at run time. Within each frame, the corresponding sub-network undergoes a pruning-while-training procedure consisting of a forward (solid line) and a backward (dashed line) pass, with $L$ as the training loss driving the learning procedure, and $s_i$ as the sparsity constraint. Referring to the example in the picture, the four frames processed in sequence are iterated for a fixed number of training steps.
The weights pruned within each frame to reach the desired sparsity $s_i$ no longer contribute in the following stages, neither to the forward nor to the backward propagation; this is illustrated in Fig.~\ref{fig:grad_mask} with the shadowed gray regions. For instance, the gradient computation from the sub-network with sparsity $s_2$, i.e., $g^{(s_2)}$, does not interfere with the previously computed gradients, i.e., $g^{(s_1)}$. This allows the entire weight-set $\theta$ to evolve during the pruning-while-training process, while ensuring that each sparse sub-network is learned considering its own gradient contribution. 
The effect of the {\em gradient masking} is twofold: first, it allows less sparse (and possibly more accurate) sub-networks to influence the weights of the more sparse and weaker ones; second, it shields the more sparse (and hence less accurate) sub-networks, preventing abrupt changes in the learning curve.

The three nested weight-sets $\{\theta^{(s_1)}$, $\theta^{(s_2)}$, $\theta^{(s_3)}\}$, which are all contained in the whole weight-set $\theta$, get isolated, processed, and returned in the form of a set of binary masks $M=\{M^{(s_1)}, M^{(s_2)}, M^{(s_3)}\}$, with $\theta^{(s_i)} = \theta \circ M^{(s_i)}$\footnote{$\circ$ indicates the Hadamard product between two matrices.}. This formulation can be generalized to any generic number of sub-networks $N$ of increasing sparsity $s_i$, resulting into a set of $N$ binary masks $M=\{ M^{(s_1)}, ..., M^{(s_N)} \}$, and thus $N$ weights subsets $\theta^{(s_i)}$.
Each mask $M^{s_i}$ is obtained through a magnitude-based rank and prune procedure over the weight-set $\theta$. Weights with lower magnitude are pruned first, until reaching the desired sparsity $s_i$ while enforcing the nesting of all weight-sets $\theta_i$:
\begin{equation}\label{eq:nested}
   s_1 < ... < s_N \textrm{ }\Rightarrow\textrm{ } \theta^{(s_1)} \supset  ... \supset \theta^{(s_N)}
\end{equation}

\begin{figure}[t]
    \centering        
    {\includegraphics[width=0.95\columnwidth]{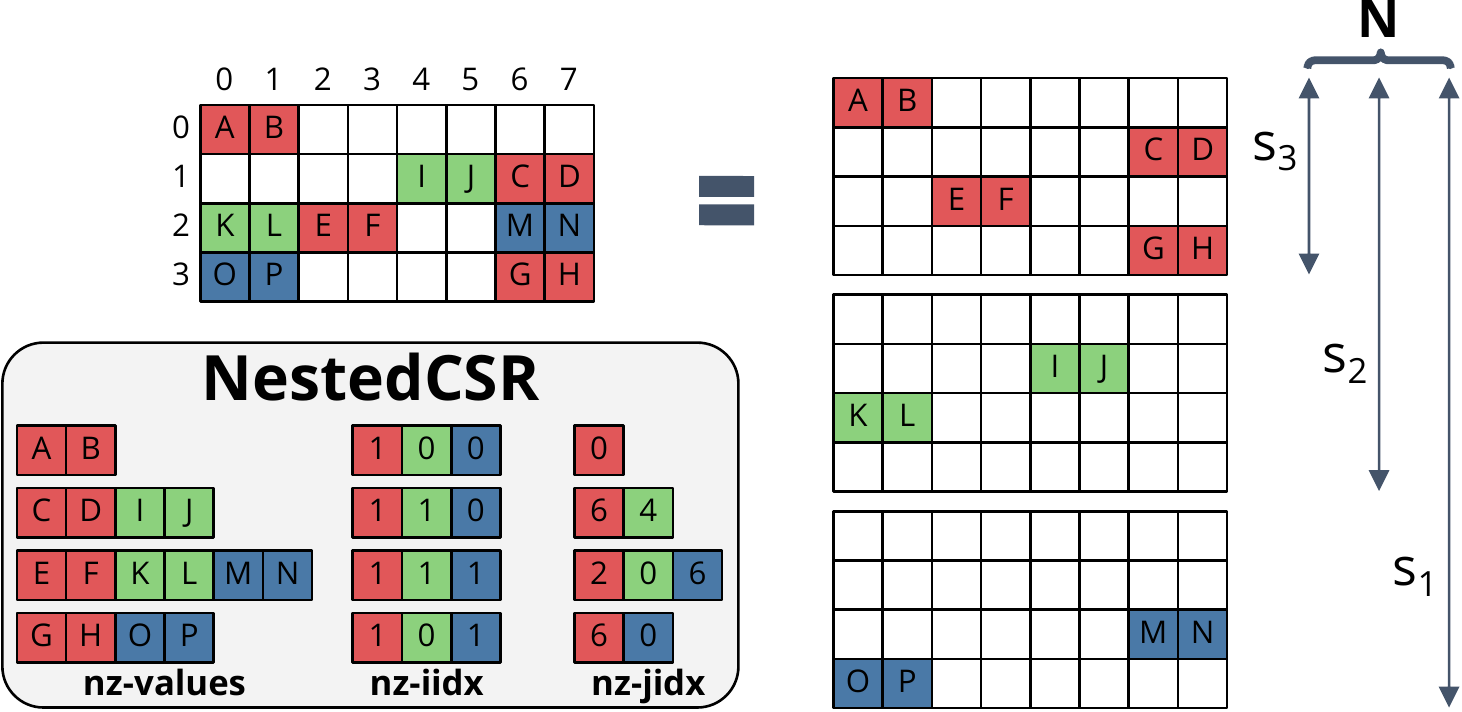}}
    \vspace{2mm}
    \caption{
        Example of the proposed NestedCSR format applied to a $1{\times}2$ block sparse matrix $W$ that can work in three sparsity levels $\{s_1, s_2, s_3\}$. \label{fig:csr}}
\end{figure}

The pseudo-code of the training loop is reported in Algorithm~\ref{alg:training}.
It takes as inputs the set of sparsity levels $S~=~\{ s_1, ..., s_N \}$ and the $block\_shape$ ($m{\times}n$), returning the weight-set $\theta$ and the set of masks $M=\{ M^{(s_1)}, ..., M^{(s_N)} \}$. 
The training loop alternates dense and sparse training epochs, according to a fixed scheduler (line~6). At the beginning of each epoch, the gradient is zeroed (line~3), then the forward and backward passes are performed on the weight-set $\theta$ (lines~4-5) as a whole (the $Dense$ training frame in Fig.\ref{fig:grad_mask}). The set of weights is directly updated using the gradient value (line~15) during the dense steps.
During the sparse training steps (the $s_i$ frames in Fig.\ref{fig:grad_mask}), for each sparsity level $s$ (line~7), the \texttt{getMask} function generates a mask $M^s$ (line~8). This mask is multiplied point-wise with $\theta$ to extract the sparse sub-network $\theta^s$ (line~9) and then used to complete the forward and backward passes (lines~10-11). For the sparse sub-networks, the predictions of the dense model (line~4) are used as soft labels  (line~10) as a form of in-place distillation~\cite{uni_slim}. At last, the local gradient $\hat{g^s}$ relative to the sub-network $\theta^s$ is masked and merged with the previous gradient contributions (line~12). 
Once the contributions of each sub-network are accumulated in the global gradient $\hat{G}$, the weight-set $\theta$ is updated (line 15).
At the end of the training, both the weight-set $\theta$ and the set of nested masks $M$ are returned (lines~17-18).
The \texttt{getMask} function used to obtain the binary mask $M^{(s_i)}$ under a given sparsity value $s_i$ works as follows. First, weights are grouped into blocks of shape $m{\times}n$ (line~21), where $m$ is in the output-channels axis. Second, blocks are ranked according to their magnitude ($L^2$-norm) through the \texttt{rankBlocks} function that returns the position ($idx$) of the sorted weights in descending order (line~22). Third, the least important $s_i \cdot |\theta|$ weights are pruned by setting to zero their values and the corresponding items in the binary mask $M^{(s_i)}$ (lines~23-24).

\subsection{Compression}\label{sec:csr}
Fig.~\ref{fig:csr} illustrates an example of the proposed sparse matrix compression format, named \textit{NestedCSR}, for a nested model trained for three generic sparsity levels $s_1<s_2<s_3$ and using a $1\times 2$ block shape. It is worth emphasizing that the compression format is general and can be used for any number of sparsity levels or block sizes.
At the lower sparsity level ($s_1$), the matrix comprises the red, green, and blue non-zero blocks; at the medium sparsity level ($s_2$), the red and green blocks; at high sparsity level ($s_3$), the red blocks only. As shown in the picture, the three configurations are a composition of three disjoint sparse matrices, and this is precisely the property exploited by \textit{NestedCSR}. 
Each sparse sub-set is compressed using a block CSR format~\cite{scalpel}: the \textit{nz-values} array stores the values of the non-zero blocks in row-major order, the \textit{nz-iidx} array stores the number of non-zero blocks on each row, and the \textit{nz-jidx} the column position of each non-zero blocks. The three arrays of each sparse sub-set are concatenated row-wise, from the most sparse to the least sparse (from red to blue in  Fig.~\ref{fig:csr}). 
\begin{figure*}[t]
    \center
    \includegraphics[width=0.95\textwidth]{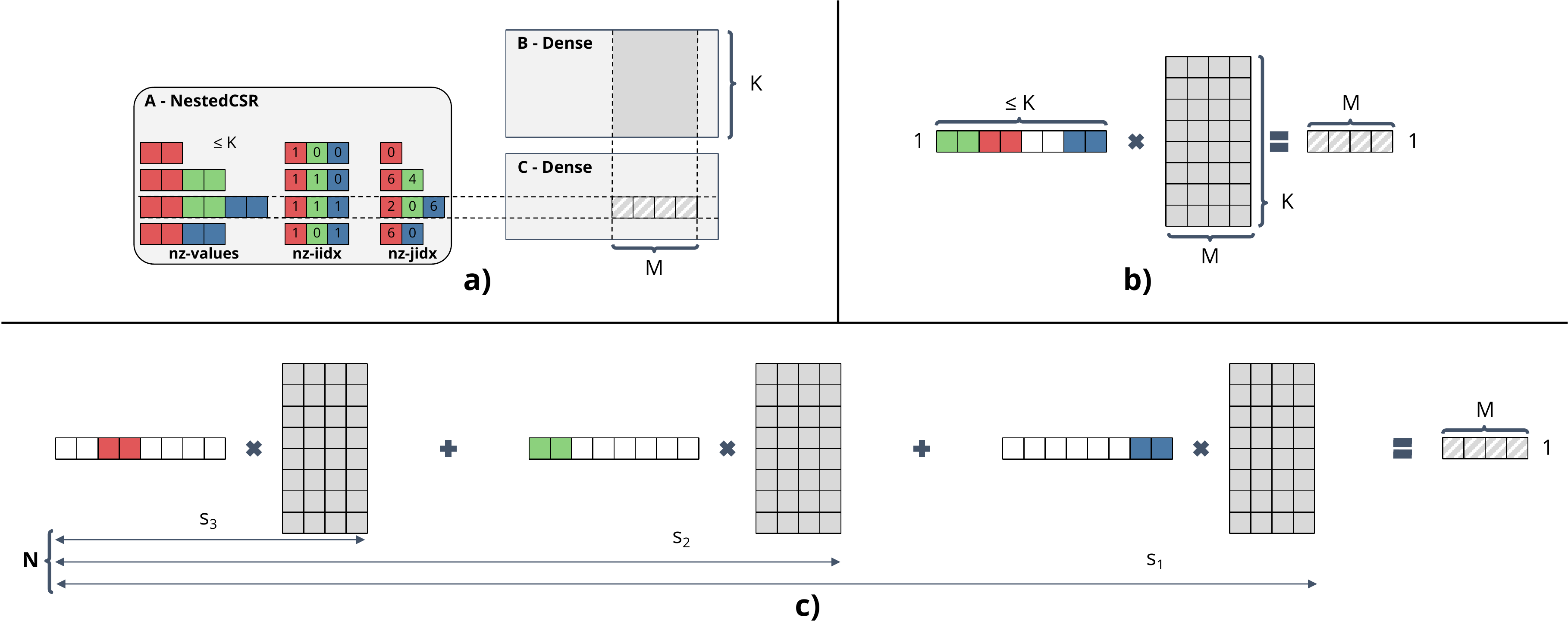}
    \vspace{2mm}
    \caption{
        Example of the proposed compute kernel performing a sparse matrix-matrix multiplication (a) between a  $1{\times}2$ block sparse matrix $A$ encoded using the NestedCSR format and a dense matrix $B$ with $K$ rows. The entire matrix multiplication is decomposed as a sequence of smaller operations (b) between $1$ row of $A$ and $M$ columns of $B$. Such inner operation is carried out as at most $N$ operations (c) depending on the selected sparsity value $s_i \in \{s_1, s_2, s_3\}$. 
    }
    \label{fig:csr_kernel}
\end{figure*}

The footprint of a block-sparse matrix $W$ with dimensions $R\times C$ encoded through \textit{NestedCSR} depends on the block shape $m{\times}n$ and the number of sparsity levels ($N$). The following equation describes the size of the array:
\begin{equation}
    \begin{aligned}
    \label{eq:nested_arrays}
        &|\textit{nz-values}| &&= (1{-}s_{min}) \cdot R {\cdot} C \\
        &|\textit{nz-iidx}| &&= N \cdot R \\
        &|\textit{nz-jidx}| &&= (1{-}s_{min}) \cdot \frac{R {\cdot} C}{n{\cdot}m}
    \end{aligned}
\end{equation}
As can be inferred from the equations, the amount of storage memory is weakly affected by the number of nested configurations. The number of sparse sub-networks ($N$) affects the size of \textit{nz-iidx}, which is usually negligible compared to that of the other two arrays. Therefore, the overall memory footprint gets defined by the smallest adopted sparsity value ($s_{min}$), which is crucial for effective and efficient deployment. 

To accelerate the processing of a nested and compressed sparse layer on a general-purpose core, we implemented a custom compute kernel that performs a matrix multiplication $C = A \cdot B$ between a sparse matrix ($A$) encoded using the \textit{NestedCSR} format and a dense matrix ($B$), as shown in fig.~\ref{fig:csr_kernel}a. 
The kernel handles both fully connected and convolutional layers, adopting a convolution-as-GEMM implementation for the convolutional layers~\cite{scalpel, lai2018cmsis}. 

Like in classical CSR-based sparse matrix multiplication, the whole operation is a sequence of small matrix operations between $M$ columns of the dense matrix and 1 row of the sparse matrix as shown in Fig.~\ref{fig:csr_kernel}b. Such implementation reduces the cost of the indirection process needed to access one element of the sparse matrix across multiple multiply-and-accumulate (MAC) operations. Specifically, it was experimentally found out that $M{=}4$ represents a good trade-off between data-reuse and register pressure on small MCUs. Following the NestedCSR format, since a single row of the sparse matrix is encoded as $N$ sparse components, the multiplication is decomposed as $N$ sparse operations at most, as shown in fig~\ref{fig:csr_kernel}c. Depending on the sparsity value $s_i$ selected at run time, only a fraction of operations is processed, exploiting the model sparsity as a practical knob to reduce the overall compute workload. In this implementation, there is no additional cost from switching the sparsity level, as the kernel can be specialized at compile time and then called at run time based on the input $s_i$ of the procedure.
\section{Results}~\label{sec:results}
\vspace{-1cm}
\subsection{Experimental Set-up}
\subsubsection{Tasks, Datasets, and ConvNets}
The proposed pipeline was tested and assessed on image classification (IC) and object detection (OD) tasks using the following data-sets.
\paragraph{CIFAR-10/100 (IC) \cite{krizhevsky2009learning}} $60k$ $32\times 32$ RGB images annotated with $10$/$100$ labels and split into $45k$ samples for training, $5k$ for validation, and $10k$ for testing.
\paragraph{PASCAL VOC (OD) \cite{pascalvoc}} $15870$ RGB images picked from the $2007$ and $2012$ PASCAL Visual Object Classes Challenge, counting of $37813$ objects annotated with $20$ different labels. As suggested in~\cite{ssd}, VOC07 and VOC12 {\em trainval} data were used for training, using VOC07 for testing. We reduced the number of classes to the top-10 labels recognized by the full-scale model. The image resolution was re-scaled to $160\times160$ with a bi-linear interpolation; this is mandatory due to the strict memory constraints of the target MCU ($512KB$ of RAM, $2MB$ of FLASH).
\begin{table}[t]
    \caption{Accuracy results for MobileNetV1 on CIFAR-10. Best results for each sparsity level are highlighted in bold.\label{tab:acc_mobv1}}
    \vspace{2mm}

    \setlength{\tabcolsep}{2pt}
    \renewcommand{\arraystretch}{1.1}
    \centering
    \resizebox{.893\columnwidth}{!}{%
    \begin{tabular}{c|c||c|c|c|c}
    \toprule
    \multirow{2}{*}{\textbf{Training}}& \textbf{Sparsity} &\multicolumn{4}{c}{\textbf{Accuracy Top-1 [\%]}}\\\cline{3-6}
    & {\bf [\%]} & {\bf w=1.00} & {\bf w=0.75} & {\bf w=0.50} & {\bf w=0.25} \\
    \midrule 
    Dense & 0 & 90.08 & 89.35 & 88.32 & 85.31 \\ 
    \midrule    
    \multirow{3}{*}{Single Sparse}
      & 70 & 89.70          & \textbf{88.56} & 87.27 & \textbf{83.32} \\
      & 80 & 89.02          & 88.13          & \textbf{87.04} & 73.22 \\
      & 90 & \textbf{88.81} & 86.02          & 75.20 & 57.88 \\
    \midrule
    \multirow{3}{*}{DSNN~\cite{wu2021dynamic}}
      & 70 & 86.30 & 86.21 & 84.09 & 78.84 \\
      & 80 & 86.42 & 85.96 & 83.69 & 76.10 \\
      & 90 & 85.49 & 84.62 & 81.78 & 72.22 \\
    \midrule 
    \midrule 
    \multirow{3}{*}{\shortstack[c]{Ours}}
      & 70 & \textbf{89.90} & 88.48          & \textbf{87.55} & 83.29 \\
      & 80 & \textbf{89.20} & \textbf{88.24} & 86.95 & \textbf{82.12} \\
      & 90 & 88.50          & \textbf{87.03} & \textbf{85.86} & \textbf{78.20} \\   
    \bottomrule    
    \end{tabular}}
    \vspace*{2mm}
\end{table}

The ConvNets used as benchmarks are lightweight models suitable for the IoT segment and hence portable onto tiny cores. Specifically, we operated ResNet (ResNet9)~\cite{resnet} for IC on CIFAR-100,  MobileNetV1\cite{howard2017mobilenets} for IC on CIFAR-10, MobileNetV2 \cite{sandler2018mobilenetv2} as backbone of the Single Shot Detector (SSD)~\cite{ssd}.

\subsubsection{Training}
The training procedure for the IC tasks was driven by the SGD optimizer (momentum $0.9$, weight decay $0.0005$) for $300$ epochs with batch size $128$. The learning rate followed a cosine annealing schedule starting from $0.05$. The same procedure applied for training the SSD, except for the batch size which was set to $32$. Images were flipped and rotated for data augmentation on the IC tasks, whereas we replicated the strategy presented in~\cite{ssd} for OD. Each training experiment was repeated three times using different seeds, and the collected results were averaged. 
For what concerns the sparse networks, we used $S$=$\{70\%$, $80\%$,$90\%\}$ as the sparsity set and a constant block shape $1\times2$ for each sparsity. Finding the optimal set $S$ to achieve the best accuracy, latency, and storage trade-off is out of the scope of this work. As suggested by previous works on sparse networks~\cite{elsen2020fast}, the first layer of each ConvNet under test is kept dense.

The training algorithm was implemented within the PyTorch framework ($v1.5.1$) and accelerated with a single consumer graphic card by NVIDIA (Titan Xp).

In the remaining sections we refer to {\em Dense} as the dense baseline network, {\em Single Sparse} as the model optimized for a single sparsity level~\cite{han2015learning}, \textit{Nested Sparse ConvNets} for our proposal, \textit{Slimmable} as the dynamic model obtained by layers width scaling~\cite{slimmable}, and {\em DSNN} as the dynamic sparse model~\cite{wu2021dynamic}. For \textit{Slimmable} we adopted the official repository\footnote{https://github.com/JiahuiYu/slimmable\_networks}, whereas for {\em DSNN} we used an in-house implementation as no open-source code was available at the time of this writing.
\begin{table}[t]
    \caption{Accuracy results for ResNet9 on CIFAR-100. Best results for each sparsity level are highlighted in bold.\label{tab:acc_cifar100_resnet9}}
    \vspace{2mm}

    \setlength{\tabcolsep}{2pt}
    \renewcommand{\arraystretch}{1.1}
    \centering
    \resizebox{.89\columnwidth}{!}{%
    \begin{tabular}{c|c||c|c|c|c}
    \toprule
    \multirow{2}{*}{\textbf{Training}}& \textbf{Sparsity} &\multicolumn{4}{c}{\textbf{Accuracy Top-1 [\%]}}\\\cline{3-6}
    & {\bf[\%]} & {\bf w=1.00} & {\bf w=0.75} & {\bf w=0.50} & {\bf w=0.25} \\
    \midrule 
    Dense & 0 & 73.78 & 72.24 & 69.66 & 63.05 \\ 
    \midrule
    \multirow{3}{*}{Single Sparse}
      & 70 & 72.93 & 71.09 & 68.29 & \textbf{58.90} \\
      & 80 & 72.61 & 70.90 & 67.72 & \textbf{57.40} \\
      & 90 & \textbf{72.15} & \textbf{69.98} & 65.04 & 52.15 \\
    \midrule 
    \multirow{3}{*}{\shortstack[c]{DSNN~\cite{wu2021dynamic}}}
      & 70 & 72.9 & 70.48 & 63.38 & 45.25 \\
      & 80 & 72.83 & 69.70 & 62.48 & 44.69 \\
      & 90 & 71.62 & 67.56 & 60.15 & 40.92 \\
    \midrule    \midrule 
    \multirow{3}{*}{\shortstack[c]{Ours}}
       & 70 & \textbf{73.56} & \textbf{72.04} & \textbf{68.82} & 58.70 \\ 
       & 80 & \textbf{72.94} & \textbf{71.05} & \textbf{68.38} & 57.30\\
       & 90 & 71.19 & 69.59 & \textbf{65.92} & \textbf{52.93} \\            
    \bottomrule    
    \end{tabular}}
    \vspace*{2mm}
    
\end{table}

\subsubsection{Deployment}
The collected performances refer to an off-the-shelf NUCLEO-F767ZI board powered by an ARMCortex-M7 MCU operating at $216MHz$. The board hosts 512KB of on-chip SRAM and $2MB$ of FLASH. An in-house extension of the CMSIS-NN library v.5.6.0~\cite{lai2018cmsis} was integrated with the sparse matrix multiplication kernels described in the previous section, with a $1\times2$ block-shape to exploit the Single Instruction Multiple Data media accelerator of the M7 core~\cite{scalpel}. In compliance with the arithmetic requirements of the CMSIS-NN library, the ConvNets were quantized to 8-bit using a layer-wise symmetric binary scaling~\cite{Jacob_2018_CVPR}. We adopted the GNU Arm Embedded Toolchain (version 6.3.1) for cross-compilation.

\subsection{Training Evaluation}
To assess the quality and generalization properties of the proposed nested training, we analyzed the accuracy achieved over the IC tasks by ConvNet architectures of decreasing information capacity, that is, rescaled by means of the width multiplier factor $w\in~\{1.00, 0.75, 0.50, 0.25\}$. Such a scaling operation must not be confused with the dynamic width scaling of~\cite{slimmable}, which is discussed later in Section~\ref{sec:lq_scaling}. The results are collected in Tab.~\ref{tab:acc_mobv1} and Tab.~\ref{tab:acc_cifar100_resnet9}.

{\bf Nested Sparse vs. Single Sparse Training.} Intuitively, training a network for a single sparsity level should be a best-case scenario because the parameters get optimized for one specific sparsity level only. On the other hand, training a Nested Sparse ConvNet encompasses the concurrent optimization of multiple sub-networks with shared weights. Nonetheless, Nested Sparse ConvNets outperform individually trained sparse models in many cases, and when they achieve a lower accuracy, the gap is rather low: the worst-case accuracy drop is $0.31\%$ for MobileNetV1 and $0.96\%$ for ResNet9. 
The gradient masking technique attains high accuracy indeed, even when classical single sparsity pruning does not. For instance, the single sparse MobileNetV1@$w{=}0.25$ with $s{=}90\%$ suffers from a drastic accuracy drop ($57.88\%$), whereas the Nested Sparse model is $20.32\%$ more accurate ($78.20\%$), closing the gap with the least sparse configurations ($83.29\%$ with $s{=}70\%$). The gradient masking technique also improves the least sparse instances due to the proper involvement of the dense model in the training loop. This can be inferred from the results collected on the Nested Sparse ResNet9@$w{=}0.75$ with $s{=}70\%$, which shows $\approx 1\%$ more accurate than its single sparse model counterpart, hence closer to the dense model.
\begin{figure*}[t]
    \centering
    \subfloat[\label{fig:latency_mobv1} MobileNetV1- CIFAR10 ]
    {\includegraphics[width=.45\linewidth]{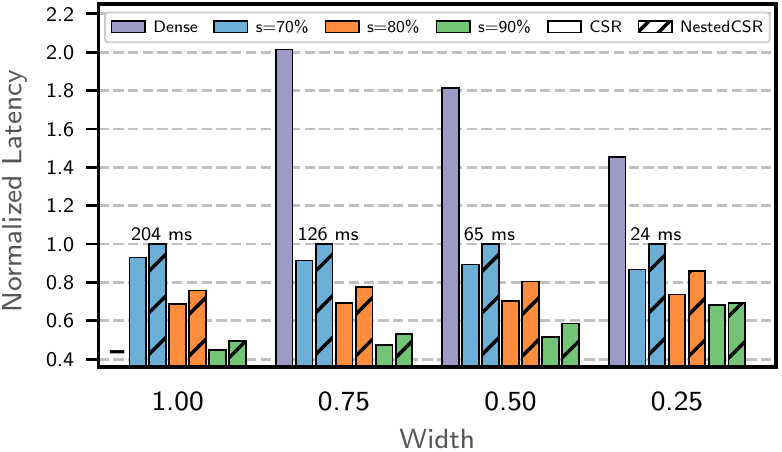}}
    \qquad
    \subfloat[\label{fig:latency_resnet9} ResNet9 - CIFAR100]
    {\includegraphics[width=0.45\linewidth]{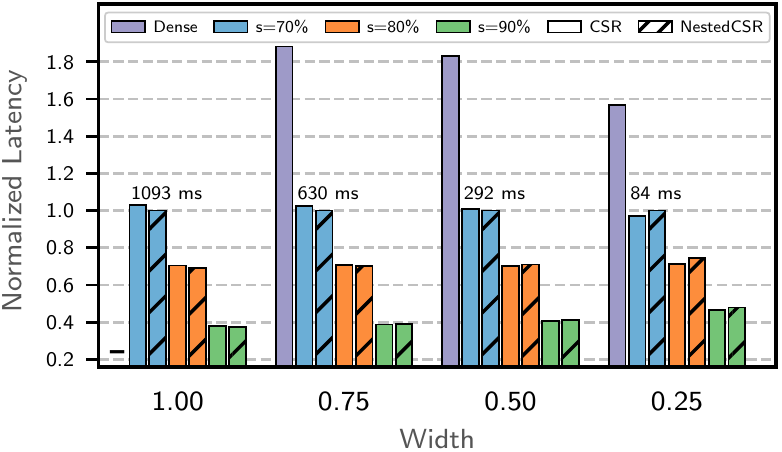}}
    \vspace{2mm}
    \caption{Latency values normalized for each width to the NestedCSR@$s{=}70\%$. The latency of the dense model at $w{=}1.00$ is not shown as it exceeds the FLASH memory of the adopted device (2MB).}
    \label{fig:latency_analysis}
\end{figure*}
\begin{table}[t]
    \caption{Storage footprint of ResNet9 trained on Cifar100 and MobileNetV1 trained on CIFAR10. {\em Single} sparse models encoded with a block CSR~\cite{scalpel}. {\em Nested} sparse models encoded with the proposed block NestedCSR format. \label{tab:fused}}
    \vspace{2mm}
    
    \setlength{\tabcolsep}{2pt}
    \renewcommand{\arraystretch}{1.1}
    \centering
   \resizebox{.98\columnwidth}{!}{%
    \begin{tabular}{c||c|c||cccc}
    \toprule
    \multirow{2}{*}{{\bf Model}} &
    \multirow{2}{*}{{\bf Method}} &
    {\bf Sparsity} & 
    \multicolumn{4}{c}{\bf Storage [KB]} \\
    & & {\bf [\%]} & {\bf w=1.00} & {\bf w=0.75} & {\bf w=0.50} & {\bf w=0.25} \\ 
    \midrule 
    & Dense  & 0                & 3132 & 1774 & 800 & 208 \\
    MobileNetV1 & Single & 70               & 1458 & 834  & 384 & 106 \\
    & Nested & $\{70, 80, 90\}$ & 1464 & 839  & 387 & 108 \\
    \midrule 
    & Dense  & 0                & 2232 & 1259 & 562 & 143 \\
    ResNet9 & Single & 70               & 1014 & 575  & 260 & 68  \\ 
    & Nested & $\{70, 80, 90\}$ & 1016 & 576  & 260 & 68 \\
    \bottomrule    
    \end{tabular}}
\end{table}

{\bf Nested Sparse vs. Dynamic Sparse NN (DSNN)}
Even though training DSNNs has proven effective on RNNs for ASR~\cite{wu2021dynamic}, our results reveal quality drops on tiny ConvNets for IC tasks. The DSNN training on MobileNetV1 is $3.40\%$ less accurate than the single sparse configuration and $13.65\%$ less on the ResNet9. Except for ResNet9@$w{=}1.00$ with $s{=}90\%$, Nested Sparse ConvNets outperform DSNNs, with an increasing gap for smaller networks with lower width and higher sparsity (the highest gap is for ResNet9@$w{=}0.25$ with $s{=}90\%$). 

\subsection{Encoding Format Evaluation}
Tab.~\ref{tab:fused} reports the storage profiles for ResNet9 and MobileNetV1, showing that Nested Sparse ConvNets achieve remarkable savings. Three nested sparse configurations require as low as $1016kB$ ($54\%$ smaller than the dense baseline) for ResNet9@$w{=}1.00$, and $1464kB$ ($53\%$ smaller) for MobileNetV1@w$1.00$. Interestingly, a Nested Sparse ConvNet takes almost the same storage of its least sparse configuration. For instance, encoding a single instance with sparsity $70\%$ using block CSR~\cite{scalpel} takes $1014KB$ for ResNet9@w=$1.00$ (a mere $2kB$ less than NestedCSR) and $1458kB$ for MobileNetV1@$w{=}1.00$ ($6kB$ less than NestedCSR). The models rescaled to the other widths follow the same trend, confirming the effectiveness of the NestedCSR format across a wide set of model configurations.

The performance attainable with the NestedCSR format further improved with the aid of the custom-designed compute kernels. Fig.~\ref{fig:latency_analysis} reports a comparative analysis for ResNet9 and MobileNetV1, both dense and sparse versions, using a classical CSR~\cite{scalpel} and the proposed NestedCSR. The sparse kernels introduce a substantial speed-up compared to the dense versions as expected, but even more remarkable, they make Nested Sparse ConvNets reach comparable performance to single sparse ConvNets. Referring to ResNet9, nested kernels perform slightly better than single sparse kernels ($1.83\%$ on average) for high widths ($w{=}1.00$ and $w{=}0.75$), and show some overhead for low width ($4.04\%$ in the worst case). For MobileNetV1, the nested kernels perform moderately worse ($10.91\%$ slower on average) and the overhead increases more notably for more sparse and smaller networks (up to $14.08\%$ in the worst case). The different internal structure of ResNet9 and MobileNetV1 is the source of such gap. In MobileNetV1, there are many convolutional layers, but only the $1\times 1$ point-wise layers are sparsified, whereas in ResNet9, there are fewer convolutional layers, but they are all sparse and also show more channels with larger kernels ($3 \times 3$). 
Despite those penalties, nested kernels still preserve the latency gain brought by sparsity. Moreover, a naive implementation of multiple sparse networks stored as separate instances would not fit on the device due to the memory constraints, an issue we overcome by means of our nested solution.  

\begin{figure*}[ht]
    \centering
    \subfloat[\label{fig:lq_scaling_mobv1} MobileNetV1- CIFAR10 ]
    {\includegraphics[width=0.41\linewidth]{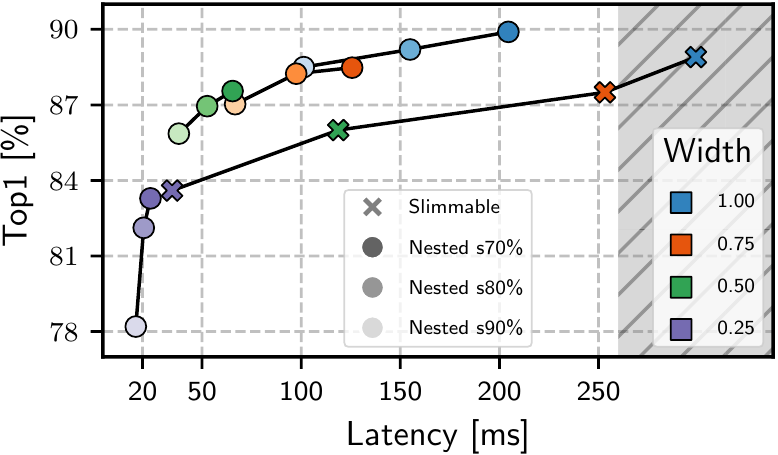}}
    \qquad
    \subfloat[\label{fig:lq_scaling_res9} ResNet9 - CIFAR100]
    {\includegraphics[width=0.41\linewidth]{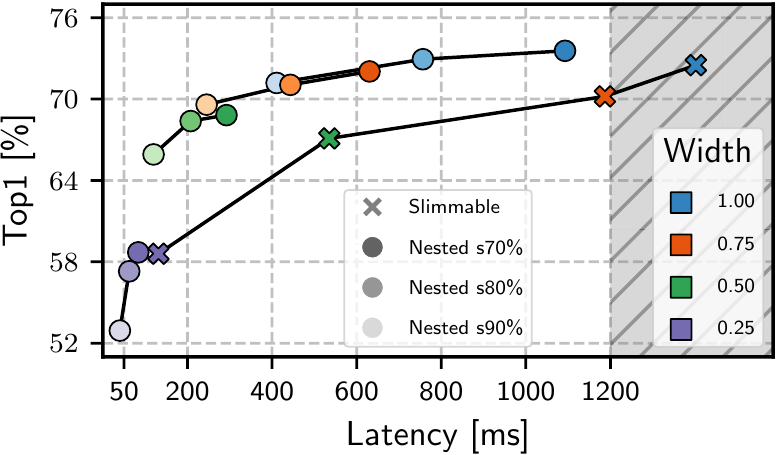}}
    \vspace{2mm}
    \caption{Latency-accuracy scaling for Slimmable ConvNets and Nested Sparse ConvNets.
    Grey area shows the unfeasible solution space for the adopted MCU, i.e., FLASH footprint $>2MB$.}
    \label{fig:lq_scaling}
\end{figure*}

\subsection{Latency-Quality Scaling}
\label{sec:lq_scaling}
Fig.~\ref{fig:lq_scaling} depicts the latency vs. accuracy trade-off achievable by Nested Sparse ConvNets. The best dynamic behavior is for larger widths. Looking at MobileNetV1@$w{=}1.00$, an increase of sparsity from $70\%$ to $90\%$ has minimal effect on accuracy ($1.4\%$), but the speed-up is substantial: up to $51\%$ of latency reduction. 
ResNet9@$w{=}1.00$ follows the same trend (Fig.~\ref{fig:lq_scaling_res9}), where a higher sparsity level improves latency by $62\%$ with a moderate effect on accuracy ($2.37\%$ loss). 
Rescaling the model width makes the trade-off slightly worse as smaller ConvNets are less resilient to sparsity. As a result, the accuracy gap increases and the latency speed-up reduces when the ConvNets architecture shrinks down. Nonetheless, for the smaller nets ($w{=}0.25$), the accuracy drop of $5.09\%$ for ResNet9 and $5.77\%$ for MobileNetV1 come with a large speed-up, $52\%$ and $31\%$ respectively.

Fig.~\ref{fig:lq_scaling} also shows the dynamic behavior of ConvNets optimized with the {\em Slimmable} approach~\cite{slimmable} offering a direct comparison with our approach. Slimmable networks at maximum width $w{=}1.00$ get too large to fit into the FLASH memory ($2MB$), and only three configurations out of four can be deployed on-device. Thanks to the proposed training and compression pipeline instead, Nested Sparse ConvNets meet the memory constraint even at full scale ($w{=}1.00$). Except for the smallest width ($w{=}0.25$), Nested Sparse ConvNets at $s{=}70\%$ and $s{=}80\%$ turn out to be more accurate and faster than the slimmable models. The {\em Pareto analysis} reveals that the three rescaled Nested Sparse ConvNets ($w{=}\{0.75, 0.50, 0.25\}$) outperform the slimmable counterparts, originating eight Pareto optimal implementations that, if stored together, consume less storage than a slimmable model. Precisely, $904kB$ for ResNet9 and $1334kB$ for MobileNetV1, that is, $28\%$ and $25\%$ less than the deployable configurations of the \textit{slimmable} models ($w\leq 0.75$). The downside is that a single Nested Sparse ConvNet presents a moderate scaling capacity compared to a slimmable model, which is intuitive as the sparsity acts as a fine-grain control knob both on accuracy and latency. However, the low storage footprint paves the way to an attractive hybrid solution, where the width multiplier serves as a static knob complementary to the dynamic sparsity.

It is worth emphasizing that other scalable training methods, e.g., \textit{EfficientNet}~\cite{tan2019efficientnet}, \textit{TinyNet}~\cite{han2020model}, and \textit{OFA}~\cite{cai2019once}, play statically, i.e., at {\em design time} , on the topology of the model architecture (i.e., width, depth, kernel sizes) and the input resolution with the aim to achieve a higher accuracy with the same resource budget. Such scaling methods are of utter importance to the design of efficient ConvNets, but their purpose differs from ours. We demonstrated that tweaking at {\em run time} the accuracy-latency trade-off via sparsity is feasible even with a reduced storage footprint, as only one compressed weight-set must be stored on-device for a Nested Sparse ConvNet. Alternatively, our solution can be used on top of existing neural architectures. 

\subsection{Object Detection}
This last subsection aims to show the generalization capability of our approach on tasks different from image classification. We evaluated a Nested MobileNetV2 on a bounding-box detection task. 
The results reported in Tab.~\ref{tab:acc_objdect} refer to configurations at $w{=}\{0.50, 0.35\}$, which are those meeting the FLASH memory constraint for our target MCU.  
The Nested Sparse object-detector gets more accurate than the sparse models trained as separate instances. For the most sparse configurations (i.e., $s{=}90\%$), it is $31.85\%$ more accurate (average over the two widths), confirming the stability of the proposed training loop. With regard to the latency, the conclusions brought by the image classification tasks do hold here, with sparse models faster than the dense models and our nested solution slightly slower than single sparse instances. Also in this case, a hybrid solution build through a superimposition of width scaling and nested sparsity does enable a wider latency-accuracy spectrum ($\Delta \textrm{Top-1}/\Delta L=12.46(\%)/368(ms)$) while cumulatively occupying $848kB$, which is still less than the single dense model at $w{=}0.50$.

\begin{table}[t]
    \caption{SSD-MobileNetV2. Best results for each sparsity level are highlighted in bold. \label{tab:acc_objdect}}
    \vspace{2mm}

    \setlength{\tabcolsep}{2pt}
    \renewcommand{\arraystretch}{1.1}
    \centering
    \resizebox{.95\columnwidth}{!}{%
    \begin{tabular}{c|c|ccc|ccc}
    \toprule
    \multirow{3}{*}{\textbf{Training}}  && \multicolumn{3}{c|}{{\bf w=0.50}} & \multicolumn{3}{c}{{\bf w=0.35}} \\ \cline{3-8}
    
    & {\bf Sparsity} & {\bf mAP} & {\bf Storage} & {\bf Latency} &  {\bf mAP} & {\bf Storage} & {\bf Latency} \\
     & {\bf [\%]} & {\bf [\%]} & {\bf [kB]} & {\bf [ms]} & {\bf [\%]} & {\bf [kB]} & {\bf [ms]}\\     
    \midrule 
    Dense & 0 & 68.32 & 869 & 1549 & 63.42 & 523 & 998 \\
    \midrule    
    \multirow{3}{*}{Single Sparse}
      & 70 & 66.01 & 508 & 1080 & 60.58 & 329 & 752 \\
      & 80 & 62.72 & 407 & 972  & 55.20 & 274 & 689  \\
      & 90 & 29.40 & 306 & 862  & 23.06 & 219 & 625 \\
    \midrule \midrule
     \multirow{3}{*}{Ours}
      & 70 &\textbf{68.30}& \multirow{3}{*}{514} & 1225 &\textbf{63.12}&\multirow{3}{*}{334} & 883  \\
      & 80 &\textbf{66.37}&                   &  1103                 &\textbf{61.03} &                  &    807               \\
      & 90 &\textbf{60.33}&                   & 951                  & \textbf{55.84}&                  & 712                   \\
    \bottomrule    
    \end{tabular}}
    \vspace*{2mm}
\end{table}

\section{Current Limitations and Future Works}~\label{sec:limitations}
\hspace{-1mm}The proposed training and compression pipeline enables the use of model sparsity as a dynamic knob on tiny off-the-shelf devices.
Although the experimental assessment revealed that Nested Sparse ConvNets outperform other dynamic strategies while occupying a smaller storage footprint, some issues have not been addressed in the current version of the work. 
First, the choice of the sparsity levels is fixed manually prior to training. However, as the trade-off accuracy vs. latency enabled by sparsity depends on the model architecture and the task, designing the optimal set of sparsity values is not trivial and should be automated.
Second, although using the same sparsity ratio for all layers of the network was proven effective in previous works~\cite{elsen2020fast}, exploiting the effects that different layers have on both accuracy~\cite{zhang2019all, molchanov2017variational} and latency~\cite{gordon2018morphnet} may lead to new Pareto solutions. 
Thus, a possible future development aimed at overcoming such limitations can integrate an automatic search engine (like those presented in~\cite{yu2019autoslim, chin2021joslim}) in the proposed pipeline such that multiple sparse configurations are sampled and tested at training time to optimize storage, latency, and accuracy simultaneously.
\section{Conclusions}~\label{sec:conlusions}
\hspace{-1mm}Nested Sparse ConvNets represent a novel class of dynamic models conceived to trade-off latency with accuracy at run time leveraging sparsity as a knob. We introduced a novel training procedure capable of reaching highly accurate predictions, and in conjunction with a new storage format and a library of custom compute kernel it enables the deployment of elastic ConvNets on tiny off-the-shelf devices. 
An extensive experimental assessment on tiny visual computing tasks deployed on a low-end node powered by an ARM M7 MCU reveals that Nested Sparse ConvNets can be processed efficiently, they outperform state-of-the-art dynamic strategies achieving optimality in the accuracy-latency objective space, and can thereby represent a new alternative for expanding the adoption of energy-efficient adaptable computer vision tasks at the edge of the IoT.

\bibliographystyle{IEEEtran}
\bibliography{refs}

\end{document}